\documentclass[runningheads]{llncs}
\usepackage{graphicx}
\usepackage{color}
\usepackage{amsmath}

\usepackage{multirow}
\usepackage{booktabs} 

\begin{document}

\iftrue
\newcommand{\alertJW}[1]{{\color{magenta}{\bf JW:#1}}}
\newcommand{\JZ}[1]{{\color{blue}{\bf JZ:#1}}}
\newcommand{\BR}[1]{{\color{red}{\bf Bog:#1}}}
\else
\newcommand{\alertJW}[1]{}
\newcommand{\JZ}[1]{}
\newcommand{\BR}[1]{}
\fi

\newcommand{\minisection}[1]{\vspace{0.04in} \noindent {\bf #1}}

\title{When Deep Learners Change Their Mind: Learning Dynamics for Active Learning\thanks{We acknowledge the support of the Spanish Ministry of Science and Innovation for funding projects PID2019-104174GB-I00.}}
%
%
\author{Javad Zolfaghari Bengar\inst{1,2}\orcidID{0000-0002-2502-8419} \and \\ 
Bogdan Raducanu\inst{1,2}\orcidID{0000-0003-2207-6260}
\and \\ 
Joost van de Weijer\inst{1,2}\orcidID{0000-0001-9843-3143}}
\authorrunning{Javad Zolfaghari Bengar et al.}
\titlerunning{Learning Dynamics for Active Learning}
%
\institute{Computer Vision Center (CVC) \and Univ. Aut\`{o}noma of Barcelona (UAB)\\
\email{\{jzolfaghari,bogdan,joost\}@cvc.uab.es}}
\maketitle              

\begin{abstract}
Active learning aims to select samples to be annotated that yield the largest performance improvement for the learning algorithm. Many methods approach this problem by measuring the informativeness of samples and do this based on the certainty of the network predictions for samples. However, it is well-known that neural networks are overly confident about their prediction and are therefore an untrustworthy source to assess sample informativeness. In this paper, we propose a new informativeness-based active learning method. Our measure is derived from the learning dynamics of a neural network. More precisely we track the label assignment of the unlabeled data pool during the training of the algorithm. We capture the learning dynamics with a metric called label-dispersion, which is low when the network consistently assigns the same label to the sample during the training of the network and high when the assigned label changes frequently. We show that label-dispersion is a promising predictor of the uncertainty of the network, and show on two benchmark datasets that an active learning algorithm based on label-dispersion obtains excellent results.
\end{abstract}

\section{Introduction}
Deep learning methods obtain excellent results for many tasks where large annotated dataset are available~\cite{krizhevsky2012learning}. However, collecting annotations is both time and labor expensive. Active Learning(AL) methods~\cite{settles2012active} aim to tackle this problem  by  reducing  the  required  annotation  effort. The  key idea behind active learning is that a machine learning model can achieve a satisfactory performance with a subset of the training samples if it is allowed to choose which samples to label. 
In AL, the model is trained on a small initial set of labeled data called initial label pool. An acquisition function selects the samples to be annotated by an external oracle. 
The newly labeled samples are added to the labeled pool and the model is retrained on the updated training set. This process is repeated until the labeling budget is exhausted. 

One of the main groups of approaches for active learning use the network uncertainty, as contained in its prediction, to select data for labelling  \cite{chitta2018largescale,wang2016cost,settles2012active}. However, it is known that neural networks are overly confident about their predictions; making wrong predictions with high certainty~\cite{ovadia2019can}. In this paper, we present a new approach to active learning. Our method is based on recent work of Toneva et al.~\cite{toneva2018an}, who study the learning dynamics during the training process of a neural network. They track for each training sample the transitions from being classified correctly to incorrectly (or vice-versa) over the course of learning. Based on these learning dynamics, they characterize a sample of being 'forgettable' (if its class label changes from subsequent presentation) or 'unforgettable' (if the class label assigned is consistent during subsequent presentations). Their method is only applicable for labeled data (and therefore not applicable to active learning) and was applied to show that redundant (forgettable) training data could be removed without hurting network performance. 

Inspired by this work, we propose a new uncertainty-based active learning method which is based on the learning dynamics of a neural network. With learning dynamics, we refer to the variations in the predictions of the neural network during training. Specifically, we keep track of the model predictions on every unlabeled sample during the various epochs of training. Based on the variations of the predicted label of samples, we propose a new active learning metric called \emph{label-dispersion}. This way, we can indirectly estimate the uncertainty of the model based on the unlabeled samples. We will directly use this metric as the acquisition function to select the samples to be labeled in the active learning cycles. Other than the forgetfulness measure proposed in~\cite{toneva2018an}, we do not require any label information. 

Experimental results show that label-dispersion better resemble the true uncertainty of the neural networks, i.e. samples with low dispersion were found to have a correct label prediction, whereas those with high dispersion often had a wrong prediction. Furthermore, in experiments on two standard datasets (CIFAR 10 and CIFAR 100) we show that our method outperforms the state-of-the-art methods in active learning. 

\section{Related work}
The most important aspect for an active learner is the strategy used to query the next sample to be annotated. 
These strategies have been successfully applied to a series of traditional computer vision tasks, such as image classification \cite{snoek2015transfer,wu2018kdd}, object detection \cite{aghdam2019active,bengar2019temporal}, image retrieval \cite{zhang2010retrieval}, remote sensing \cite{deng2018pr}, and regression \cite{denzler2018bmvc}.

Pool based methods are grouped into three main query strategies relying mostly on heuristics: informativeness
\cite{hauptmann2015ijcv,gal2017icml,gu2014modelchange}, representativeness \cite{sener2018active}, and
hybrid \cite{zhou2014hybrid,loog2018maxvariance}, 
a comprehensive survey of these frameworks and a detailed discussion can be found in \cite{settles2012active}. 

\minisection{Informativeness-based methods:}
Among all the aforementioned strategies, the informativeness-based approaches are the most successful ones, with uncertainty being the most used selection criteria in both bayesian \cite{gal2017icml} and non-bayesian frameworks \cite{hauptmann2015ijcv}. In \cite{Yoo_2019_CVPR,li2020learning}, the authors employed a loss module to learn the loss of a target model and select the images based on their output loss. More recently, query-synthesizing approaches have used generative models to generate informative samples \cite{mahapatra2018efficient,mayer2020adversarial,zhu2017generative}. 

\minisection{Representativeness-based methods:}
In \cite{Sinha_2019_ICCV} the authors rely on selecting few examples by increasing diversity in a given batch. The Core-set technique was shown to be an effective representation learning method for large scale image classification tasks \cite{sener2018active} and was theoretically proven to work best when the number of classes is small. However, as the number of classes grows, its performance deteriorates. Moreover, for high-dimensional data, using distance-based representation methods, like Core-set, is ineffective because in high-dimensions p-norms suffer from the curse of dimensionality which is referred to as the distance concentration phenomenon in the computational learning literature \cite{donoho2000high}.

\minisection{Hybrid methods:}
Methods that aim to combine uncertainty and representativeness use a two-step process to select the points with high uncertainty as of the most representative points in a batch \cite{li2013adaptive}.  
A weakly supervised learning strategy was introduced in \cite{wang2016cost} that trains the model with pseudo labels obtained for instances with high confidence in predictions. While most of the hybrid approaches are based on a two-step process, in \cite{Wang2020daal} they propose a method to select the samples in a single step, based on a generative adversarial framework. An image selector acts as an acquisition function to find a subset of representative samples which also have high uncertainty. 
\begin{figure*}[t]
\centering
\includegraphics[width=1\columnwidth]{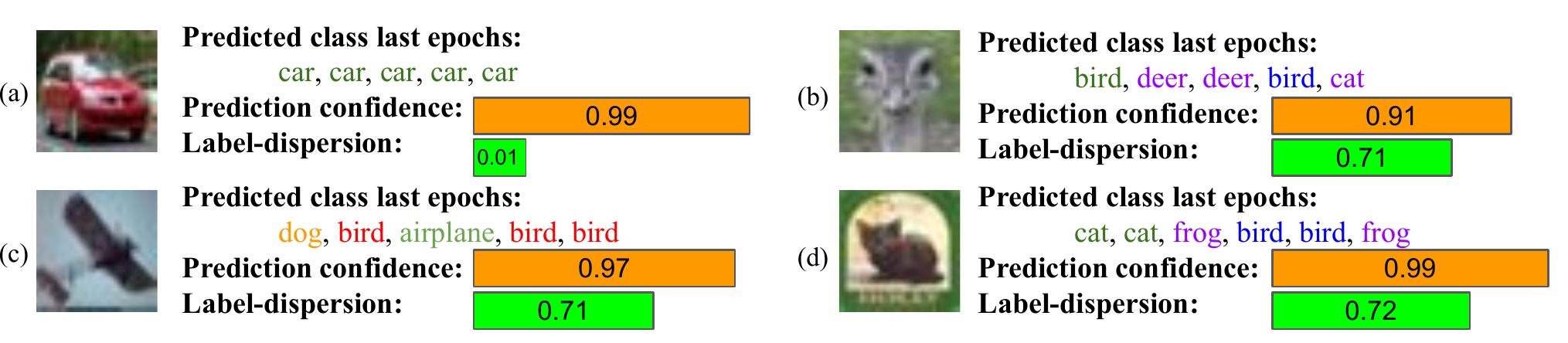}
    \caption{\small\textbf{Comparison between the dispersion and confidence scores.} 
    We show four examples images together with the predicted label for the last five epochs of training. The last predicted label is the network prediction when training is finished. We also report the prediction confidence and our label-dispersion measure. (a) Shows an example which is consistently and correctly classified as \emph{car}. The confidence of model is 0.99 and the consistent predictions every epoch result in low dispersion score of 0.01. (b-d) present examples on which the model is highly confident despite a wrong final prediction and constant changes of predictions across the last epochs. This network uncertainty is much better reflected by the high label-dispersion scores.}
\label{fig:conf_vs_disp}
\end{figure*}

\section{Active learning for image classification}

We describe here the general process of active learning for the image classification task. Given a large pool of unlabeled data $U$ and an annotation budget $B$, the goal of active learning is to select a subset of $B$ samples to be annotated as to maximize the performance of an image classification model. Active learning methods generally proceed sequentially by splitting the budget in several cycles. Here we consider the batch-mode variant \cite{sener2018active}, which annotates multiple samples per cycle, since this is the only feasible option for CNN training. At the beginning of each cycle, the model is trained on the initial labeled set of samples. After training, the model is used to select a new set of samples to be annotated at the end of the cycle via an acquisition function.   The selected samples are added to the labeled set $\mathcal{D}_L$ for the next cycle and the process is repeated until the total annotation budget is spent. 

\subsection{Label-dispersion acquisition function}
In this section, we present a new acquisition function for active learning. The acquisition function is the most crucial component and the main difference between active learning methods in the literature. In general, an acquisition function receives a sample and outputs a score indicating how valuable the sample is for training the current model. Most of informativeness-based active learning approaches consider to assess the certainty of the network on the unlabeled data pool which is obtained after training on the labeled data \cite{chitta2018largescale,wang2016cost,settles2012active}. 

In contrast, we propose to track the labels of the unlabeled samples during the course of training. We hypothesize that if the network frequently changes the assigned label, it is unsure about the sample, therefore the sample is an appropriate candidate to be labeled. In figure \ref{fig:conf_vs_disp} we depict the main idea behind our method and compare it to network confidence. While the confidence score is used to assign the label based on the certainty of the last epoch, the dispersion uses the prediction over all epochs in order to assess the certainty. The first example shows the case of a correct label prediction when both confidence score and dispersion agree. However, in the other three examples, we depict situations where the system predicts the wrong label with high certainty. However, a large dispersion value (i.e. high uncertainty) is the indication of an erroneous prediction.

\begin{figure*}[t]
\centering
\includegraphics[width=1\columnwidth]{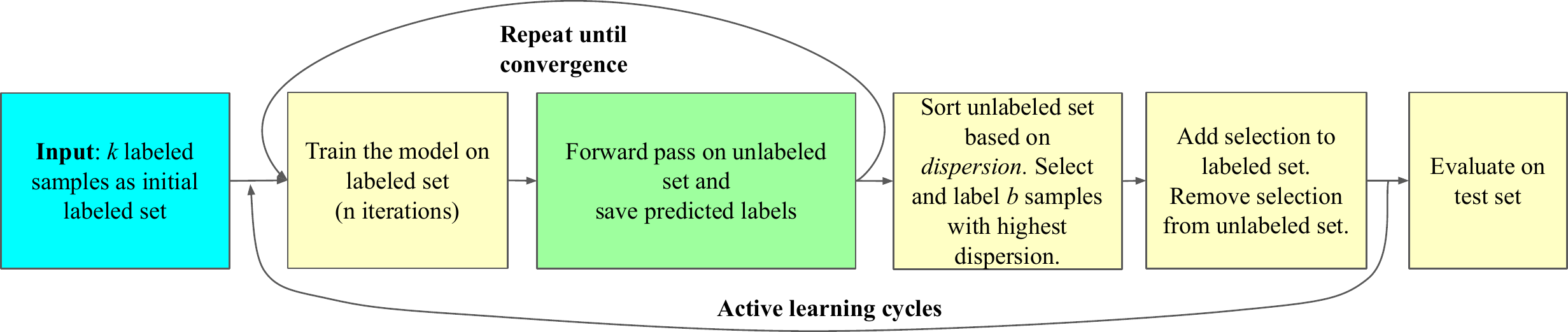}
    \caption{\small\textbf{Active learning framework using Dispersion.} Active learning cycles start with initial labeled pool. The model trained on labeled pool is used to output the predictions and compute dispersion for each sample. The samples with highest dispersion are queried for labeling and added to labeled set. This cycle repeats until the annotation budget is exhausted.}
\label{fig:framework}
\end{figure*}

This idea is based on the concept of \emph{forgettable samples} recently introduced by \cite{toneva2018an}. \cite{toneva2018an} states that there exist a large number of unforgettable samples that are never forgotten once learnt. It is shown that they can be omitted from the training set while the generalization performance is maintained. Therefore it suffices to learn the forgettable samples in the train set. However to identify forgettable samples the ground-truth labels is needed.  Since we do not have access to the labels in active learning, we propose to use a measure called the \emph{label-dispersion}. The dispersion of a nominal variable is calculated as the proportion of predicted class labels that are not the modal class prediction \cite{freeman1965elementary}. It estimates the uncertainty of the model by measuring the changes in the predicted class as following:
\begin{equation}
    Dispersion(x) := 1-\frac{f_x}{T},\label{eq_disp}
\end{equation}
with 
\begin{equation}
    \begin{split}
        f_x &= \sum_{t} 1[y^t=c^*],\\
        c^* &=\underset{c=1,...,C} {\arg\max}  \sum_{t} 1[y^t=c],
    \end{split}
\end{equation}
where $f_x$ is the number of predictions falling into the modal class for sample $x$ and $C$ is the number of classes. Larger values for dispersion means more uncertainty in model outputs. Similar to forgettable samples, we are interested in samples for which the model doesn't persistently output the same class.

Fig.~\ref{fig:framework} presents the active learning framework with our acquisition function. During the training of a network at regular intervals we will save the label predictions for all samples in the unlabeled pool (green block in Fig.~\ref{fig:framework}). In practice, we will perform this operation at every epoch. These saved label predictions allow us to compute the label-dispersion with Eq.~\ref{eq_disp}. We then select the samples with highest dispersion to be annotated and continue to the next active learning cycle until the total label budget is used.

\begin{figure*}[t]
\centering
\includegraphics[width=.9\textwidth]{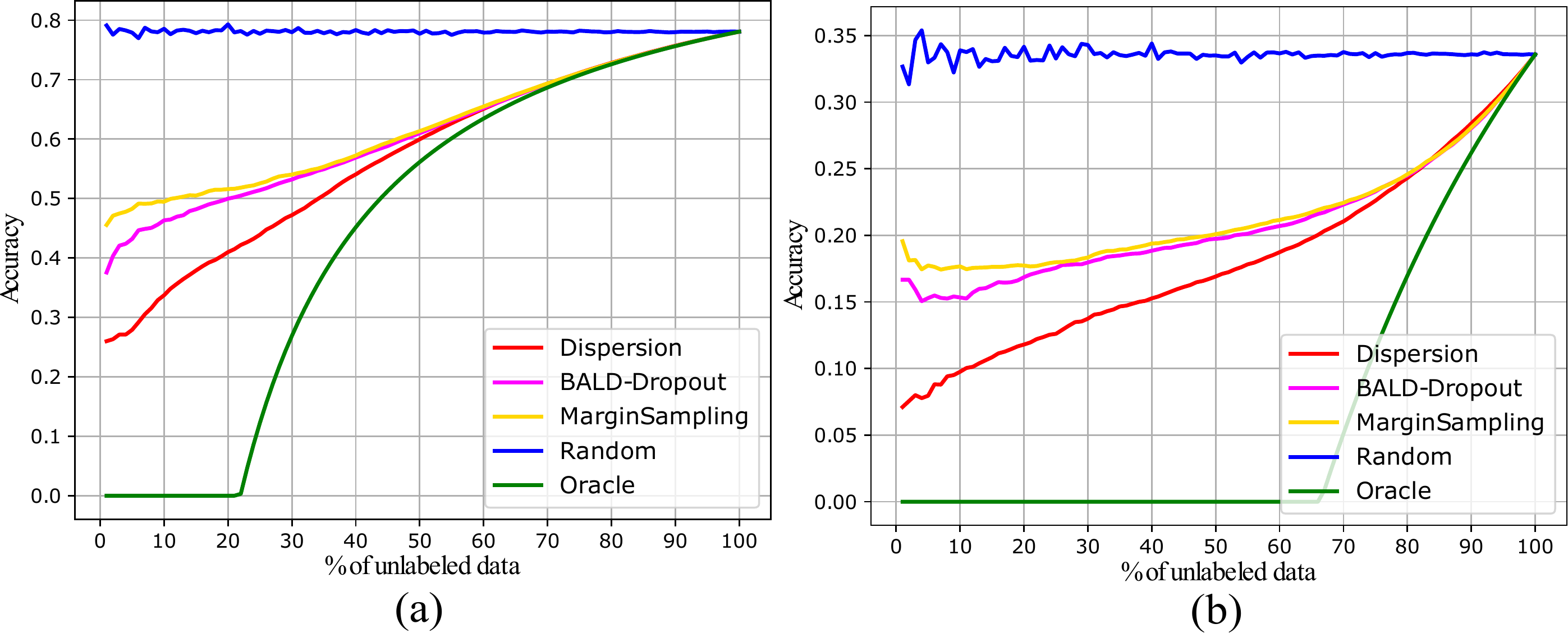}
\caption{\small \textbf{Informativeness analysis of AL methods on CIFAR10(a) and CIFAR100(b) datasets.} The model is used to infer the label of samples selected by AL methods before labeling and the accuracy is measured. For any amount of unlabeled samples, dispersion offers samples with lower accuracy and hence more informative for the model.}
\label{fig:combination_cifar_10_100}
\end{figure*}

\subsection{Informativeness Analysis}
To assess the informativeness of methods, we compute the scores assigned to the unlabeled samples and sort the samples accordingly. Then we select several portions of the most informative samples (according to their score) and run the model to infer their labels. We argue that annotating the correctly classified samples would not provide much information for the model because the model already knows their label. In contrast, the model can learn from misclassified samples if labeled.
We use the accuracy to implicitly measure the informativeness of unlabeled samples. The lower the accuracy, the more informative the samples will be if labeled. Fig.~\ref{fig:combination_cifar_10_100} shows the accuracy of model on the unlabeled samples queried by each method. The model used in this analysis is trained on the initial labeled set. The accuracy of samples selected randomly remains almost constant regardless of the amount of unlabeled samples. In this analysis, the oracle method by definition uses groundtruth and queries samples that the model misclassified and therefore the accuracy of the model is zero. Among the active learning methods, on both CIFAR10 and CIFAR100 datasets, and for any amount of unlabeled samples, dispersion queries misclassified samples the most, showing that high dispersion correlates well with network uncertainty. These samples can potentially increase the performance of the model if labeled.   

\section{Experimental Results}
\subsection{Experimental Setup}
We start with model trained on initial labeled set from scratch and employ Resnet-18 as the model architecture. The initial labeled set consists of $10\%$ of train dataset that is selected randomly once for all the methods. At each cycle, we use the model with the corresponding acquisition function to select $b$ samples, which are then labeled and added to $\mathcal{D}_L$. We continue for 4 cycles until the total budget is completely exhausted. In all experiments, the budget per cycle is 5$\%$ and total budget is 30$\%$ of the entire dataset. Eventually for each cycle, we evaluate the model on the test set. 
To evaluate our method, we use CIFAR10 and CIFAR100 \cite{krizhevsky2012learning} datasets with 50K images for training and 10K for test. CIFAR10 and CIFAR100 have 10 and 100 object categories respectively and image size of 32$\times$32. During training, we apply a standard augmentation scheme including random crop from zero-padded images, random horizontal flip, and image normalization using the channel mean and standard deviation estimated over the training set. 

Dispersion is computed from the most probable class in the output of the model. During training we do an inference on the unlabeled pool at every epoch and save the model predictions. Based on these predictions we compute the label-dispersion for each sample specifically.

\minisection{Implementation details.}
Our method is implemented in PyTorch\footnote{Upon acceptance, we will release the code for our method.} \cite{paszke2017automatic}. We trained all models with the momentum optimizer with value 0.9 and the initial learning rates 0.02. We train for 100 epochs and reduce the learning rate by a factor of 5 once after 60 epochs and again at 80 epochs. Finally, to obtain more stable results we repeat the experiments 3 times and report the mean and standard deviation in our results.
\minisection{Baselines.} We compare our method with several informative and representative-based approaches. \emph{Random sampling:} selects an arbitrary subset of samples from all unlabeled samples. \emph{BALD} \cite{gal2017icml}: method chooses samples that are expected to maximise the information gained about the model parameters. In particular, it select samples that maximise the mutual information between predictions and model posterior via dropout technique. \emph{Margin sampling}~\cite{brust2019active}: uses the difference between the two classes with the highest scores as a measure of proximity to the decision boundary. \emph{KCenterGreedy}~\cite{sener2018active}: is a greedy approximation of KCenter problem also known as min-max facility location problem \cite{wolf2011facility}. Samples having maximum distance from their nearest labeled samples in the embedding space are queried for labeling. \emph{CoreSet} \cite{sener2018active}: finds samples as the 2-Opt greedy solution of Kcenter problem in the form of Mixed Integer Programming (MIP) problem. \emph{VAAL} \cite{Sinha_2019_ICCV}: learns a latent space using a Variational Autoencoder (VAE) and an adversarial network trained to discriminate between unlabeled and labeled data. The unlabeled samples which the discriminator classifies with lowest certainty as belonging to the labeled pool are considered to be the most representative and queried for labeling. \emph{Oracle method:} An acquisition function using ground-truth that selects samples that the model miss-classified. In order to study the potential of active learning, we evaluate oracle-based acquisition function. Note this is not a useful active learning function in practice, as we would not have access to the ground-truth annotations in a real scenario.  In order to make a fair comparison with the baselines, we used their official code and adapted them into our code to ensure an identical setting.

\subsection{Results}
\begin{figure*}[t]
\includegraphics[width=1\columnwidth]{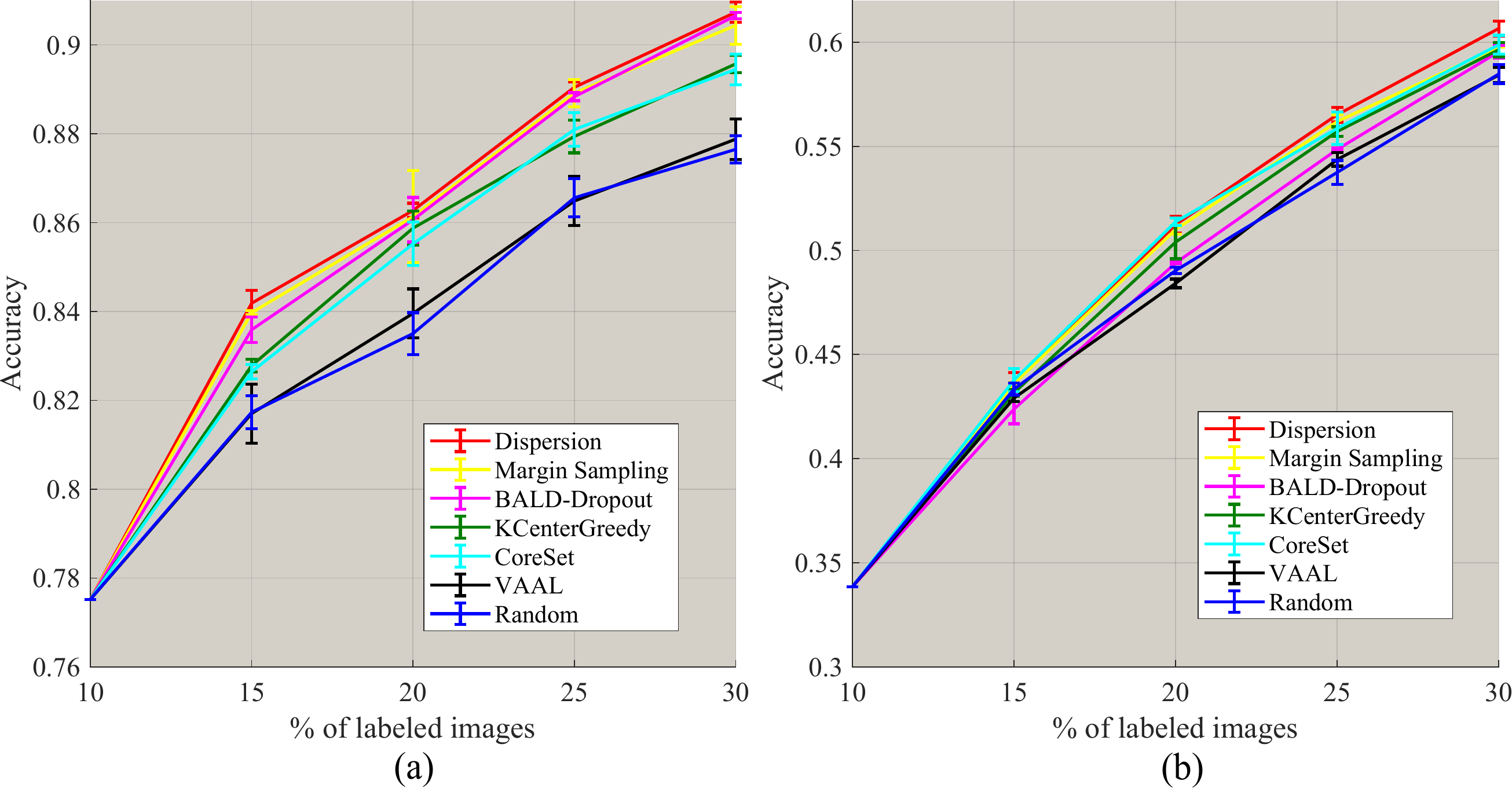}
    \caption{\small\textbf{Performance Evaluation.} Results for several active learning methods on CIFAR10 (a) and CIFAR100 (b) datasets. All curves are average of 3 runs.}
\label{fig:ALC_cifa10_cifar100}
\end{figure*}
\minisection{Results on CIFAR10:}
A comparison with several active learning methods, including both informativeness and representativeness, is provided in  Fig.~\ref{fig:ALC_cifa10_cifar100}. As can be seen in Fig.~\ref{fig:ALC_cifa10_cifar100}(a) dispersion outperforms the other methods across all the cycles on CIFAR10, only the BALD-Dropout method obtains similar results at $30\%$. The active learning gain of dispersion against Random sampling is around 7.5$\%$ at cycle 4, equivalent to annotating 4000 samples less. The informative methods such as Margin Sampling and BALD lie above the representative methods including KCenterGreedy, CoreSet, VAAL and Random highlighting the importance of informativeness on CIFAR10 where the number of classes is limited and each class is well-represented by many samples. 

\minisection{Results on CIFAR100:} Fig.~\ref{fig:ALC_cifa10_cifar100}(b) shows the performance of active learning methods on CIFAR100. As can be seen, the methods are closer and the overall performance of Dispersion, Margin sampling and CoreSet are comparable. However, the addition of labeled samples at cycle 3 and 4 makes the dispersion superior in performance to others. The smaller gap between the informative based methods and Random emphasizes the importance of representativeness on CIFAR100 dataset which has more diverse classes that are underrepresented with few samples in small budget size.    

Additionally, Table~\ref{tab:performance_wrt_full_set} illustrates the  full performance of models that are trained on the entire datasets. Dispersion manages to attain almost 97$\%$ and 82$\%$ of full performance on CIFAR10 and CIFAR100 respectively by using only $30\%$ of the data, which is a significant reduction in the labeling effort.

\begin{table}[ht]
    \setlength{\tabcolsep}{12pt}
    \caption{\small  \textbf{Active learning results.} Performance of AL methods using 30$\%$ of dataset both in absolute performance and relative to using all data.\vspace{-2mm}}
    \centering
    \begin{tabular}{l cccc}
        \toprule
         \multirow{2}{*}{Methods} & \multicolumn{2}{c}{CIFAR 10} &  \multicolumn{2}{c}{CIFAR 100} \\
         & Acc. & Rel. & Acc. & Rel. \\
         \midrule
         All data & 93.61 & 100\% & 74.61 & 100\% \\
         \midrule
         Dispersion & \textbf{90.74} & 96.93\% & \textbf{60.66} & 81.97\%\\
         Margin sampling \cite{brust2019active}  & 90.44 & 96.61\% & 59.78 & 80.78\%\\
         BALD \cite{gal2017icml} & 90.66 & 96.85\% & 59.54 & 80.46\%\\
         KCenterGreedy \cite{sener2018active} & 89.57 & 95.69\% & 59.64 & 80.59\%\\
         CoreSet \cite{sener2018active} & 89.45 & 95.56\% & 59.87 & 80.91\%\\
         VAAL \cite{Sinha_2019_ICCV} & 87.88 & 93.88\% & 58.42 & 78.95\%\\
         Random sampling & 87.65 & 93.63\% & 58.47 & 79.02\%\\
         \bottomrule
    \end{tabular}
    \label{tab:performance_wrt_full_set}
\end{table}

\section{Conclusion}
We proposed an informativeness-based active learning algorithm based on the learning dynamics of neural networks. We introduced the label-dispersion metric, which measures label-consistency during the training process. We showed that this measure obtains excellent results when used for active learning on a variety of benchmark datasets. 
For future work, we are interested in exploring label-dispersion for other research fields such as out-of-distribution detection and within the context of lifelong learning. 

\bibliographystyle{splncs04}
\bibliography{shortstrings,longstrings,egbib}
\end{document}